\documentclass[letterpaper, 10 pt, journal, twoside]{ieeetran}

\IEEEoverridecommandlockouts                              

\newtheorem{theorem}{Theorem}[section]

\newtheorem{problem}{Problem}

\usepackage{graphicx} 
\usepackage{amsmath}
\usepackage{amssymb}
\usepackage{epstopdf}
\usepackage{cite}
\usepackage[noend,ruled,linesnumbered]{algorithm2e}
\SetKwComment{Comment}{$\triangleright$\ } {}
\usepackage{multirow}
\usepackage{rotating}
\usepackage{subfigure} 
\usepackage{color} 
\usepackage{mysymbol}
\usepackage[dvipsnames]{xcolor}
\usepackage[hyphens]{url}
\usepackage{romannum}
\usepackage{makecell}

\usepackage[breaklinks=true, colorlinks, bookmarks=true, citecolor=Black, urlcolor=Violet,linkcolor=Black]{hyperref}

\usepackage{comment}
\usepackage[colorinlistoftodos,prependcaption,textwidth=1.5cm,textsize=tiny]{todonotes}
\begin{document}

\title{NAMO-LLM: Efficient Navigation Among Movable Obstacles \\with Large Language Model Guidance}
%

\markboth{IEEE Robotics and Automation Letters. Preprint Version. Accepted September, 2025}
{Zhang \MakeLowercase{\textit{et al.}}: NAMO-LLM} 

\author{Yuqing Zhang, Yiannis Kantaros
\thanks{Manuscript received: May, 6, 2025; Revised August, 18, 2025; Accepted September, 27, 2025. This paper was recommended for publication by Editor Aniket Bera upon evaluation of the Associate Editor and Reviewers' comments.
This work was supported by the NSF CAREER award CNS $\#2340417$ and the  NSF award CNS $\#2231257$.} 
\thanks{The authors are with the Department of Electrical and Systems Engineering, McKelvey School of Engineering, Washington University in St.Louis, St. Louis, MO, 63130, USA.
{\tt\footnotesize \{zyuqing, ioannisk\}@wustl.edu}}%
\thanks{Digital Object Identifier (DOI): see top of this page.}
}

\maketitle
\begin{abstract}
%
Several planners have been proposed to compute robot paths that reach desired goal regions while avoiding obstacles. However, these methods fail when all pathways to the goal are blocked. In such cases, the robot must reason about how to reconfigure the environment to access task-relevant regions—a problem known as Navigation Among Movable Objects (NAMO). While various solutions to this problem have been developed, they often struggle to scale to highly cluttered environments. To address this, we propose NAMO-LLM, a sampling-based planner that searches over robot and obstacle configurations to compute feasible plans specifying which obstacles to move, where, and in what order. Its key novelty is a non-uniform sampling strategy guided by Large Language Models (LLMs) biasing the search process toward directions more likely to yield a solution. We show that NAMO-LLM is probabilistically complete and demonstrate through experiments that it efficiently scales to cluttered environments, outperforming related works in both runtime and plan quality.

\end{abstract}


\maketitle

\section{Introduction}

\begin{figure}[t]
\centering
    \includegraphics[width=0.8\linewidth]{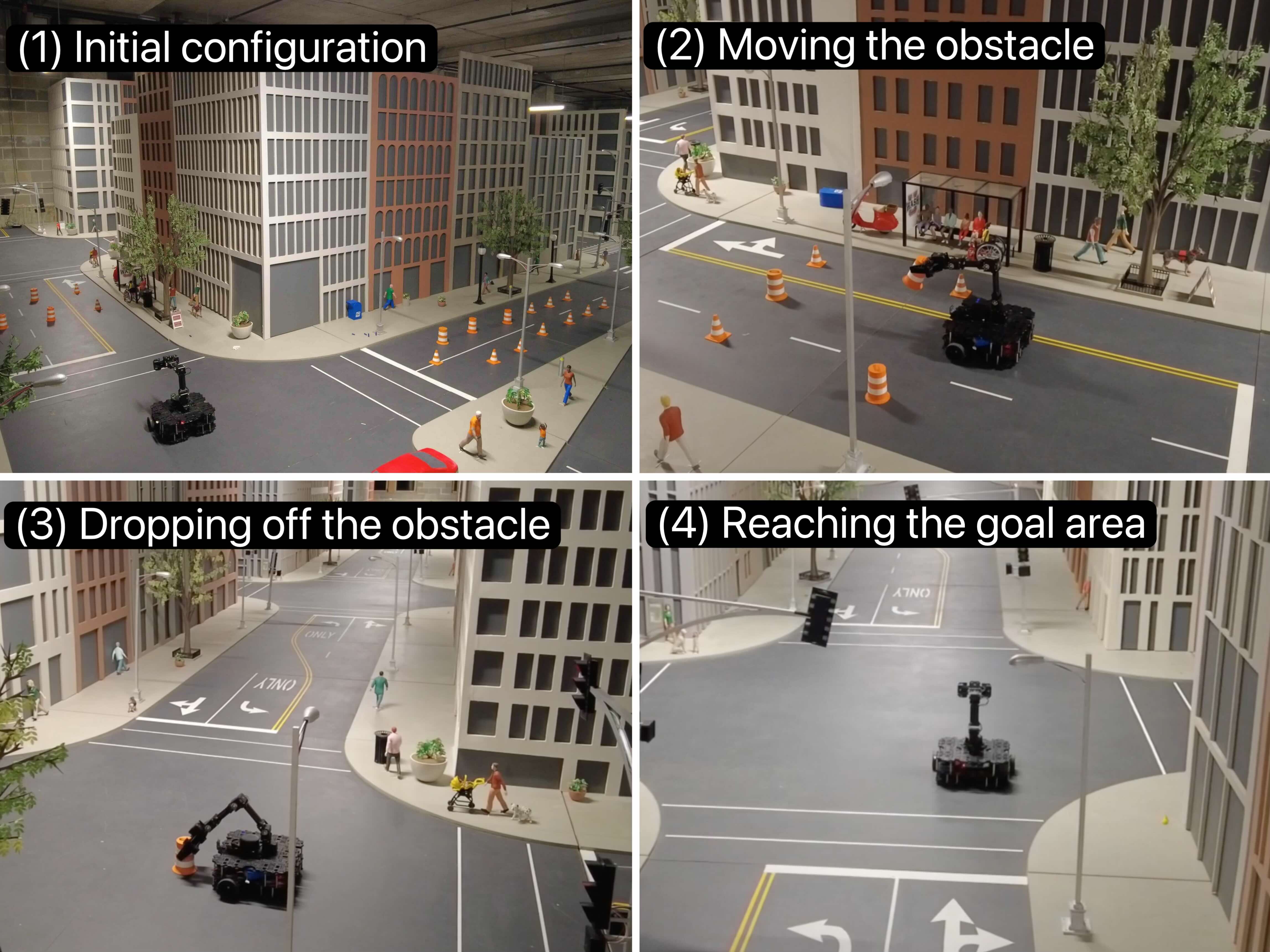} 
    \vspace{-0.1cm}
\caption{\textcolor{black}{A robot must reach a destination behind the building block. Since all direct paths are blocked, it relocates obstacles to access the goal region.}}
    \label{pf2}
    \vspace{-0.3cm}
\end{figure}

%
\IEEEPARstart{S}{everal} planning algorithms have been developed to compute robot paths from an initial configuration to a desired one while avoiding obstacles \cite{karaman2011sampling, pendleton2017numerical}.
However, in highly cluttered environments, obstacle-free paths to goal regions may not exist; see Fig. \ref{pf2}. In such cases, rather than reporting task failure, a more effective approach is to enable robots to interact with their environment by relocating movable obstacles. This requires planners capable of determining which obstacles to move, in what order, and where to reposition them to create a feasible path to the goal; see Fig. \ref{pf2}. This problem, known as Navigation Among Movable Objects (NAMO)  has been proven to be NP-hard \cite{wilfong1988motion}. 

Early approaches to address NAMO problems relied on heuristic solutions which lacked completeness guarantees \cite{chen1990practical}. To overcome this, search-based planning methods have been introduced. These methods either provide completeness guarantees for specific subclasses of NAMO problems or sacrifice computational efficiency to ensure completeness across a broader range of scenarios \cite{stilman2005navigation,stilman2007planning,stilman2008planning,moghaddam2016planning}. 
Extensions of these approaches to unknown environments are considered in 
\cite{wu2010navigation,levihn2013hierarchical,levihn2013planning,ellis2023navigation,armleder2024tactile}. 
Sampling-based planners have been developed to improve efficiency while still maintaining (probabilistic) completeness guarantees for general NAMO problems \cite{nieuwenhuisen2008effective,stilman2007manipulation,van2010path,bayraktar2023solving}. Their key idea is to incrementally construct trees by sampling nodes, where each node represents the environment’s configuration, and edges model feasible manipulation actions. However, their efficiency declines as the number of movable obstacles or the number of required obstacle relocations grows, due to their uniform sampling strategy.


\textcolor{black}{This paper addresses this limitation by designing a new sampling-based NAMO algorithm that efficiently scales to environments with many movable obstacles and/or requiring multiple obstacle relocations to reach target regions.} Inspired by \cite{van2010path}, our algorithm incrementally builds trees exploring the configuration space of movable objects and the robot’s free-space components. The key novelty lies in a non-uniform sampling strategy that leverages Large Language Models (LLMs) to accelerate plan generation. Specifically, the strategy directs the tree toward LLM-recommended directions at certain times while allowing random exploration otherwise. We refer to the resulting planner as NAMO-LLM and show it is probabilistically complete, meaning the probability of computing a feasible plan (if one exists) converges to one as the number of iterations grows. \textcolor{black}{Extensive numerical and hardware experiments demonstrate that NAMO-LLM outperforms search-based planners \cite{ellis2023navigation}, uniform-sampling planners \cite{van2010path}, and planners relying solely on LLMs in terms of runtime for the first feasible plan, the number of obstacles relocated, and task success rates}. This performance gap increases with environment complexity. 

\textbf{Related Works:} In addition to the NAMO algorithms discussed earlier, our work is also related to \textit{rearrangement planning (RP)} problems where robots manipulate movable obstacles \cite{labbe2020monte,krontiris2015dealing,11128108,ren2022rearrangement}. RP differs from NAMO because it considers predefined goal positions for the movable obstacles and typically does not specify a final goal for the robot. We believe that our non-uniform sampling strategy can also be integrated with various sampling-based approaches for planning among movable obstacles, including RP and manipulation planning \cite{simeon2004manipulation,schmitt2017optimal}; however, this integration is beyond the scope of this paper.
Furthermore, several works design \textit{non-uniform sampling strategies} to accelerate sampling-based planners 
\cite{luo2021abstraction,wang2020neural,ichter2018learning,liu2024nngtl,johnson2023learning}. 
However, we emphasize that these works target different planning problems from ours, focusing on designing obstacle-free paths between an initial and a desired robot configuration (while also assuming that such paths exist). As such, they cannot be directly applied to our setting.  It is worth noting that, unlike 
\cite{wang2020neural,ichter2018learning,liu2024nngtl,johnson2023learning}, our work does not require training neural networks; instead, ours leverages pre-trained LLMs.

\textbf{Summary of Contributions:} \textit{First}, \textcolor{black}{we introduce NAMO-LLM, a novel sampling-based planner designed to efficiently solve NAMO problems in highly cluttered environments.} \textit{Second}, we propose the first integration of LLMs within a sampling-based NAMO planner, significantly accelerating plan generation. \textcolor{black}{\textit{Third}, we conduct extensive experiments demonstrating that our approach scales well to complex environments and outperforms related planners in efficiency, solution quality, and task success rates.}

\section{Problem Formulation}\label{sec:problem}


\textbf{Robot Modeling:} We consider a robot modeled as: $\bbp(t+1)=\bbf(\bbp(t),\bbu(t)),$
where $\mathbf{p}(t)\in\ccalP$ and $\mathbf{u}(t)\in\ccalU$ denote the robot configuration (e.g., position and orientation) and the control input at time step $t$, respectively. We assume that $\bbp(t)$ is always known and that the robot is holonomic and, therefore, it can move along any desired path. The robot is also equipped with a gripper, enabling it to manipulate one movable object at a time.

\begin{figure}[t]
\centering
    \includegraphics[width=1\linewidth]{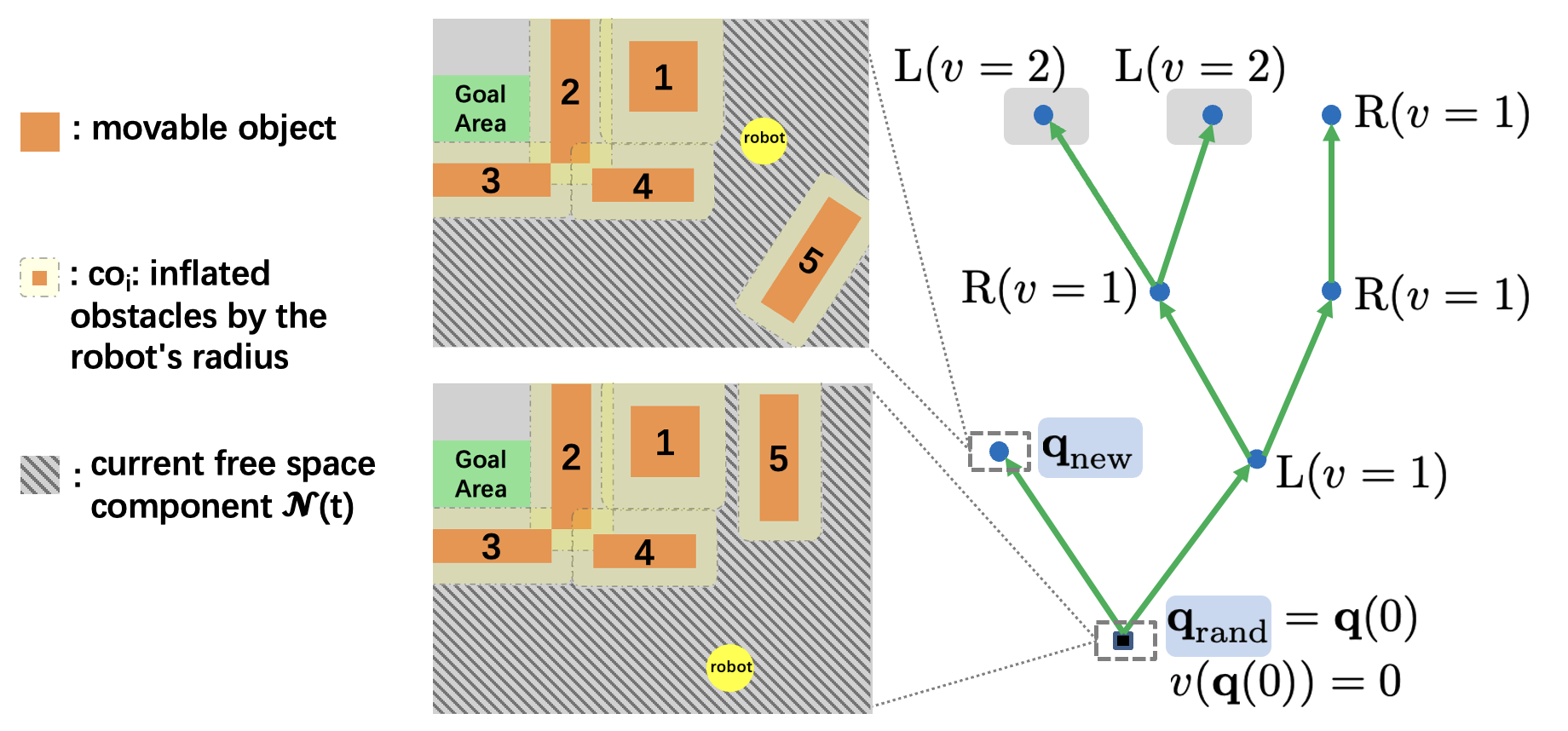}
    \vspace{-0.5cm}
\caption{\textcolor{black}{\textcolor{black}{Tree structure built by the proposed planner. Each node (blue) represents a state $\bbq(t)$, and each edge (green arrow) represents a manipulation action. For instance, $\bbq_{\text{new}}$ is reached from $\bbq_{\text{rand}}$ via action $a(\bbc_5, \bbc_5')$ moving obstacle $o_5$. If moving $o_5$ is recommended by the LLM (rather than random sampling), the score $v(\bbq_{\text{new}})$ is set to $1$; otherwise, it remains $0$. The label next to each node indicates whether it was generated by moving an LLM-suggested obstacle (L) or a randomly selected obstacle (R) (see Section \ref{M2}). The number next to the label is the score $u(\bbq)$ yielding a set $\ccalV_v$ that collects the top two gray-shaded nodes (see Section \ref{M2}). The two snapshots illustrate the environment  associated with $\bbq_{\text{rand}}$ and $\bbq_{\text{new}}$.}  
}}
\vspace{-0.2cm}
\label{fig:tree_structure}
\end{figure}

\textbf{Environment Modeling:} The robot resides in an environment $\Omega$ containing a set of known obstacles $\ccalO=\{o_1,\dots,o_O\}$. 
The obstacles are classified as movable obstacles, collected in a set $\ccalO_{\text{movable}}\subseteq \ccalO$,  and non-movable ones. 
A movable object can be repositioned by the robot as long as it first grasps it. The configuration (e.g., position and orientation) of an obstacle $o_i \in \ccalO$ at time $t$ is denoted by $c_i(t)\in\ccalC_i$ where $\ccalC_i$ denotes its configuration space. We also denote by ${\bf{co}}_i(t)$ the region in $\mathcal{P}$ that corresponds to configurations where the robot is in collision with obstacle $o_i$. Thus, the robot's free space at time $t$, denoted by $\ccalP_{\text{free}}(t)$, is defined as 
$\ccalP_{\text{free}}(t)=\ccalP\setminus \mathcal{CO}(t)$ where $\mathcal{CO}(t)=\cup_{i=1}^n\{{\bf{co}}_i(t)\}$. 
Notice that $\ccalP_{\text{free}}(t)$ may consist of multiple connected components whose boundaries are determined by the configuration of the obstacles. We denote by $\ccalN(t)\subseteq \ccalP_{\text{free}}(t)$, the free-space component that $\bbp(t)$ lies in; \textcolor{black}{see also Fig. \ref{fig:tree_structure}}.

\textbf{Robot Task:} The robot must reach a desired configuration from an initial one $\bbp(0)$ while avoiding obstacles. Let $\ccalP_g \subseteq \ccalP$ denote the desired configurations. We focus on cases where all configurations in $\ccalP_g$ are initially unreachable due to obstacle placements, i.e., $\ccalP_g \cap \ccalN(0) = \emptyset$. The robot must then determine which movable obstacles to relocate and their target placements. To formalize this, we define the set of manipulable obstacles at time $t$ as $\mathcal{O}_M(t)\subseteq \ccalO_{\text{movable}}$ which includes all movable obstacles  $o_i$ that the robot can directly reach, i.e., $\mathcal{O}_M(t)=\{o_i\in \ccalO_{\text{movable}}\mid {\bf{co}}_i(t)\cap\ccalN(t)\neq\emptyset\}$. Next, we define manipulation actions of the form $a(c_i(t),c_i')$ requiring the robot to approach, grasp, and relocate an an object $o_i\in \mathcal{O}_M(t)$, from its current configuration $c_i(t)$ to a new configuration $c_i(t')=c_i'$, for some $t'>t$. Our goal is to compute a plan $\tau = \tau(0), \dots, \tau(H)$ such that $\ccalP_g \cap \ccalN(H) \neq \emptyset$ for some horizon $H \geq 0$. Here, $\tau(k)$ denotes the $k$-th manipulation action, with $k \in \{0, 1, \dots, H\}$. Note that $k$ indexes actions, not time steps, as executing $\tau(k)$ may span multiple time steps $t$. 


\begin{problem} \label{pr1}
Given $\bbp(0)$,  $\ccalP_g$, $\ccalO$, $\ccalO_{\text{movable}}$, and $c_i(0)$ for all $o_i$, compute $\tau=\tau(0),\dots, \tau(H)$ so that $\ccalP_g\cap\ccalN(H)\neq \emptyset$.
\end{problem}

\section{Proposed Planning Algorithm}\label{sec:method}
In this section, we present NAMO-LLM, our planner for Problem \ref{pr1}. Section \ref{M1} describes a sampling-based planner for generating plans $\tau$. Section \ref{M2} introduces a non-uniform sampling strategy leveraging LLMs to accelerate plan synthesis. In Section \ref{M3}, we discuss the input prompt to the LLM guiding the sampling strategy. In Section \ref{sec:probCompl}, we show that NAMO-LLM is probabilistically complete. 

\subsection{Sampling-based Planning} \label{M1}

We propose a sampling-based planner to address Problem \ref{pr1}, summarized in Alg. \ref{alg1}.
The algorithm incrementally builds a tree exploring the configuration space of the movable objects and the robot's free-space components to compute $\tau$; \textcolor{black}{see Fig.~\ref{fig:tree_structure}}. 
We denote the constructed tree by $\ccalG=\{\ccalV,\ccalE\}$, where $\ccalV$  and $\ccalE$
is a set of nodes and edges, respectively. The set $\ccalV$ contains states of the form $\bbq(t)=  [\mathcal{N}(t), \bbc(t) ]$, where $\bbc(t)$ is a vector stacking the configuration of all obstacles, i.e., $\bbc(t)=[\bbc_1(t),\bbc_2(t),\dots,\bbc_n(t)]$ and  $\ccalN(t)$ denotes the free-space component that $\bbp(t)$ lies in. The root of the tree models the initial configuration of the obstacles and the robot, i.e., $\bbq(0)=[\ccalN(0),\bbc(0)]$ [lines \ref{alg1:free}-\ref{alg1:ini}, Alg. \ref{alg1}]. The tree is initialized so that $\ccalV=\{\bbq(0)\}$ and $\ccalE=\emptyset$ [line \ref{alg1:ini_tree}, Alg. \ref{alg1}]. Also, the tree is built incrementally by sampling new states $\bbq_{\text{new}}$ that are added to the tree structure [line \ref{alg1:pick}, Alg. \ref{alg1}]. The algorithm terminates after a user-specified number of iterations $m_{\text{max}}$ and
returns a feasible plan $\tau$ (if it has been found). In what follows, we present our method in detail.

\begin{algorithm}[t]
\footnotesize
\caption{NAMO-LLM}
\LinesNumbered
\label{alg1}
\KwIn{ (i) initial robot configuration $\bbp(0)$; (ii) initial obstacle configuration $\bbc(0)$; (iii) Set $\ccalO_{\text{movable}}$ ; (iv) goal set $\ccalX_g$ }
\KwOut{The manipulation plan $\tau$}
Compute $\ccalN(0)$\; \label{alg1:free}
$\bbq(0) = [ \ccalN(0), \bbc(0)] $ \; \label{alg1:ini}
Initialize $ \ccalV = \{\bbq_0\},~\ccalE= \emptyset,~\ccalX_g= \emptyset,~  m=1$ \;  \label{alg1:ini_tree}
\While{$m < m_{\text{max}}$}{
    Sample a node $\bbq_{\text{rand}}\in\ccalV$ using  $f_\ccalV$\; \label{alg1:pick}
    $[\bbq_{\text{new}}, a(\bbc_i(t), \bbc_i')] = \texttt{TreeExpansion}(\bbq_{\text{rand}}) $\; \label{alg1:exp}
    \If{ $\bbq_{\text{new}}\neq \varnothing$}{
    $\ccalV \gets \ccalV \cup \{\bbq_{\text{new}}\} $, $\ccalE \gets \ccalE \cup (\bbq_{\text{rand}},\bbq_{\text{new}})$\; \label{alg1:tree}
    $A((\bbq_{\text{rand}},\bbq_{\text{new}}))=a(\bbc_i(t), \bbc_i')$\; \label{alg1:storeAction}
    Extract $\ccalN(t')$ from $\bbq_{\text{new}}=[\mathcal{N}(t'), \bbc(t')]$\;
    \If{ $\ccalP_g \cap \mathcal{N}(t')\neq\emptyset$ }{ \label{alg1:if}
    $\ccalX_g\gets\ccalX_g \cup \{\bbq_{\text{new}}\}$\;\label{alg1:updXg}
    }
    }
    Update iteration index: $m=m+1$\; \label{alg1:upd-m} 
    }
    \If{$\ccalX_g\neq\emptyset$}{\label{alg1:nonemptySol}
    Pick a node $\bbq_g\in\ccalX_g$\;\label{alg1:pick-qg}
    Compute path $\tau_{\bbq}$ over $\ccalG$ from $\bbq(0)$ to $\bbq_g$\;\label{alg1:tauq}
    Using $A$ and $\tau_{\bbq}$, compute the manipulation plan $\tau$ \; \label{alg1:output} 
    \textcolor{black}{Compute robot motion path $\xi$ executing $\tau$\; \label{alg1:motion}}
    }
    \Else{
    No plan $\tau$ has been found\;\label{alg1:noplan}
    }
\end{algorithm}

\begin{algorithm}[t]
\footnotesize
\caption{\texttt{TreeExpansion}($\bbq_{\text{rand}}$)}
\LinesNumbered
\label{alg2}
Compute the set of manipulable obstacles $\ccalO_{M,\text{rand}}(t)$ \; \label{alg2:mani}
Sample a manipulable obstacle $o_i$ from a given distribution $f_{\ccalO_M}$  \; \label{alg2:picko} 
Sample a random configuration $\bbc_i'$ from $f_{\ccalC_i}$\; \label{alg2:pickc}
\If{${\bf{co}}_i'\in\ccalN^i_{\text{rand}}(t)$ \textit{and} ${\bf{co}}_i'\in\ccalP_{\text{free}}(t)\setminus {\bf{co}}_i(t)$ \label{alg2:check0}}{
\textcolor{black}{Select $\bbp(t)\in\ccalN_{\text{rand}}(t)$ and $\ccalN'$ and compute $\zeta$}\;\label{alg2:computeZeta}
\If{$\exists~\zeta$ \label{alg2:check}}{
Initialize $\bbc(t')=\bbc(t)$ and update $\bbc_i(t')=\bbc_i'$ in $\bbc(t')$\; \label{alg2:newO}
$\mathcal{P}_{\text{free}}(t') = \mathcal{P}_{\text{free}}(t) \cup {\bf{co}}_i(t) \setminus {\bf{co}}_i' $ \; \label{alg2:newfree}
$\bbq_{\text{new}}= (\mathcal{N}', \bbc(t'))$ \; \label{alg2:newq}
\Return $\bbq_{\text{new}}$, $a(\bbc_i(t),\bbc_i')$\label{alg2:act} 
}
\Else{{\label{alg2:elses}}
\Return $\bbq_{\text{new}} = \varnothing, ~a(\bbc_i(t),\bbc_i')= \varnothing$ \label{alg2:else0}}     
}
\Else{
    \Return $\bbq_{\text{new}}=\varnothing,~a(\bbc_i(t),\bbc_i')=\varnothing$ \label{alg2:else}
}
\end{algorithm}

At every iteration $m$ of Alg. \ref{alg1}, we perform the following two steps. First, we sample from a given discrete mass function $f_{\ccalV}:\ccalV\rightarrow[0,1]$ a tree node denoted by $\bbq_{\text{rand}}(t)=[\ccalN_{\text{rand}}(t),\bbc_{\text{rand}}(t)]\in\ccalV$ [line \ref{alg1:pick}, Alg. \ref{alg1}]. 
%
%
Second, we sample a new node $\bbq_{\text{new}}$ that will be added to the tree; see Alg. \ref{alg2}. Specifically, given $\bbq_{\text{rand}}$, we compute the corresponding set of manipulable obstacles denoted by $\mathcal{O}_{M,\text{rand}}(t)$, [line \ref{alg2:mani}, Alg. \ref{alg2}]. Then, we sample from a  discrete mass function $f_{\ccalO_M}:\ccalO_{M,\text{rand}}\rightarrow[0,1]$ a manipulable obstacle $o_i \in \mathcal{O}_M(t)$ [line \ref{alg2:picko}, Alg. \ref{alg2}]. 
Given the selected manipulable obstacle $o_i$, we sample a new configuration $\bbc_i'\in\ccalC_i$ using a density function $f_{\mathcal{C}_i}: \mathcal{C}_i\rightarrow [0,\infty)$ [line \ref{alg2:pickc}, Alg. \ref{alg2}]. Designing the functions $f_{\ccalV}$, $f_{\ccalO_M}$, and $f_{\ccalC_i}$ is deferred to Section \ref{M2}.

\begin{figure}
    \centering
    \includegraphics[width=\linewidth]{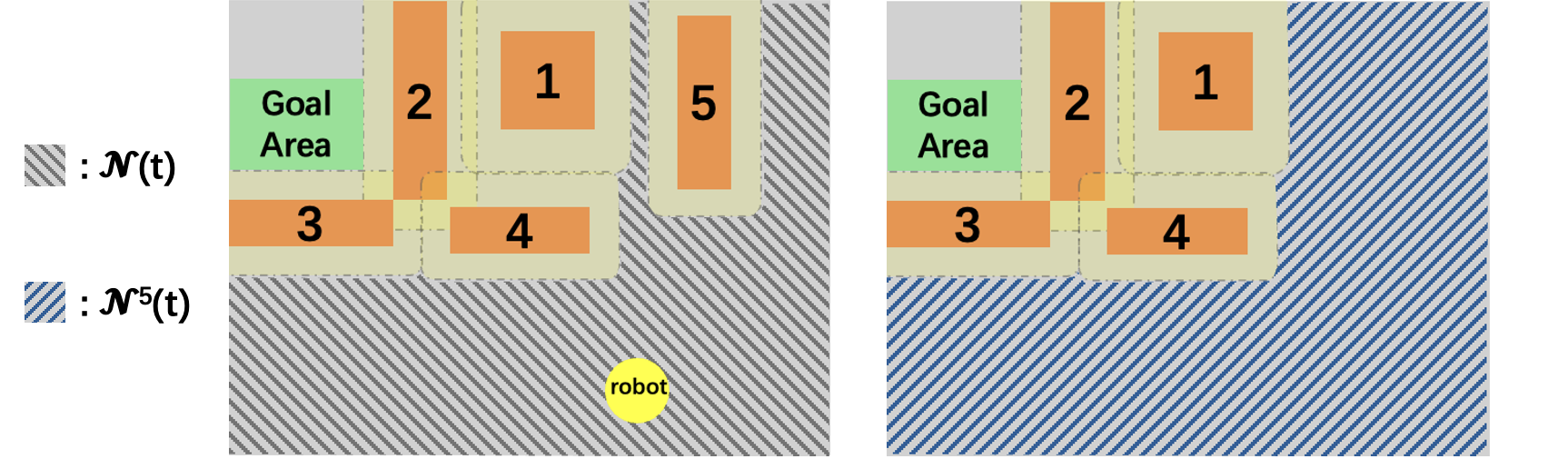}\vspace{-0.2cm}
    \caption{\textcolor{black}{Illustration of the components $\ccalN(t)$ (left) and $\ccalN^5(t)$ (right). Uniform vs non-uniform sampling for obstacle selection: In the left figure, only obstacles 1, 3, 4, and 5 are manipulable.  Under uniform sampling, each obstacle $o_i$ will be selected with probability $p_i$ defined as $p_1=p_3=p_4=p_5=0.25$ and $p_2=0$. Under non-uniform sampling, if the LLM-recommended obstacle $o_{\text{LLM}}$ is $o_3$ and $p_{\text{obs}}=0.8$, then the corresponding probabilities become $p_3=0.85$, $p_1=p_4=p_5=0.05$, and $p_2=0$.}}
    \label{fig:distribution}
    \vspace{-0.3cm}
\end{figure}

Next, we check if a collision-free path exists for the robot to execute $a(\bbc_i,\bbc_i')$, i.e., to move $o_i$ from $\bbc_i(t)$ to $\bbc_i'$ [line \ref{alg2:check}, Alg. \ref{alg2}]. \textcolor{black}{Transporting $o_i$ first requires the robot to move from its current configuration in $\ccalN_{\text{rand}}$ to ${\bf{co}}_i(t)$ in order to grasp $o_i$. Such a path always exists regardless of the exact robot configuration in $\ccalN_{\text{rand}}$ because $o_i$ is manipulable (i.e., ${\bf{co}}_i \cap \ccalN_{\text{rand}} \neq \emptyset$) and the robot is holonomic (see Section \ref{sec:problem}). Therefore, in what follows, we focus only on the existence of a collision-free path that enables the robot to move $o_i$ from $\bbc_i(t)$ to $\bbc_i'$, assuming the robot is already grasping $o_i$.} \textcolor{black}{Two necessary conditions must be met for $\bbc_i'$ to admit such a path 
[line \ref{alg2:check0}, Alg. \ref{alg2}]. First, $\bbc_i'$ must not intersect other obstacles, i.e., ${\bf{co}}_i'\in\ccalP_{\text{free}}^{\text{rand}}(t)\setminus {\bf{co}}_i(t)$, where $\ccalP_{\text{free}}^{\text{rand}}(t)$ is computed using $\bbq_{\text{rand}}$. Second, ${\bf{co}}_i'$ must not lie in a component of $\ccalP_{\text{free}}^{\text{rand}}$ that becomes accessible only if an obstacle $o_j\neq o_i$ is relocated first. To formalize this, we define $\ccalP^i_{\text{free}}(t)$ as the robot's free space if obstacle $o_i$ is removed from the environment, i.e., $\ccalP^i_{\text{free}}(t)= \ccalP_{\text{free}}(t) \setminus {\bf{co}}_i(t)$. Correspondingly, $\ccalN^i(t)$ denotes the component of $\ccalP^i_{\text{free}}(t)$ containing the robot's position $\bbp(t)$; see Fig. \ref{fig:distribution}. 
Thus, the second condition requires ${\bf{co}}_i'$ to belong to $\ccalN^i_{\text{rand}}(t)$ (i.e., to $\ccalN^i(t)$ computed using $\bbq_{\text{rand}}$).}

\textcolor{black}{If these two conditions hold, then a collision-free path to execute $a(\bbc_i,\bbc_i')$ (as soon as it grasps $o_i$) may exist. This path is defined for the robot while carrying $o_i$. Thus, with slight abuse of notation, we define this path as $\zeta = [\bbp(t),\bbc_i(t)], [\bbp(t+1),\bbc_i(t+1)], \dots, [\bbp(t+s),\bbc_i(t+s)], \dots, [\bbp(t'),\bbc_i(t')]$ where $\bbc_i(t') = \bbc_i'$, ${\bf{co}}_i(t+s) \in \ccalN^i_{\text{rand}}(t)$ for all $s\in \{1,2,\dots,t'-t\}$, and the starting configuration $\bbp(t) \in \ccalN_{\text{rand}}(t)$ is within a platform-specific threshold $\epsilon$ from $o_i$, enabling grasping. Note that if $o_i$ is moved to $\bbc_i'$, the set $\ccalN^i_{\text{rand}}(t)$ may split into multiple components, each adjacent to ${\bf{co}}_i'$. After the move, the robot may be located in any of these components; we randomly select one, denoted $\ccalN'$, and require $\bbp(t') \in \ccalN'$. Existence of $\zeta$ can be verified using any motion planner (e.g., RRT \cite{karaman2011sampling}) [line \ref{alg2:computeZeta}]. Under the considered holonomic dynamics (see Section \ref{sec:problem}), the exact starting location $\bbp(t)$ of the robot does not affect the feasibility of transporting the object.} 

If $\zeta$ exists, then we construct $\bbc(t')$ that differs from $\bbc(t)$ only in the configuration of $o_i$ [line \ref{alg2:newO}, Alg. \ref{alg2}] and we set $\ccalN(t')=\ccalN'$. Then, we construct the node $\bbq_{\text{new}}=[\ccalN(t'),\bbc(t')]$ [line \ref{alg2:newq}, Alg. \ref{alg2}] and the sets of nodes and edges are updated as $\ccalV=\ccalV \cup \{\bbq_{\text{new}}\}$ and $\ccalE=\ccalE \cup (\bbq_{\text{rand}},\bbq_{\text{new}})$ [lines \ref{alg1:tree}, Alg. \ref{alg1}].\footnote{\textcolor{black}{We note that Alg. \ref{alg1}–\ref{alg2} do not specify the exact time step $t'$ at which the node $\bbq_{\text{new}}$ is reached. The time stamp is included only to indicate that $\bbq_{\text{new}}$ will be reached at some future step. The actual time step $t$ at which a tree node $\bbq$ is reached is determined by the robot’s motion path $\xi$ used to execute a plan $\tau$ that passes through $\bbq$; computation of $\xi$ is discussed later. }} Also, once an edge $(\bbq_{\text{rand}},\bbq_{\text{new}})$ is added to the tree, we store the corresponding manipulation $a(\bbc_i(t),\bbc_i')$ that can enable it\textcolor{black}{; see Fig. \ref{fig:tree_structure}.} We formalize this by iteratively constructing a function $A$ which returns the corresponding action for each edge in $\ccalE$ [line \ref{alg1:storeAction}, Alg. \ref{alg1}]. If a collision-free path $\zeta$ does not exist or if $\bbc_i'$ is not a valid configuration, then we set $q_{\text{new}}=\varnothing$ and the iteration $m$ will terminate without adding any new nodes to the tree [lines \ref{alg2:elses}-\ref{alg2:else}, Alg. \ref{alg2}].

Once a node $\bbq_{\text{new}}$ is added to the tree, we check if $\ccalP_g\cap \ccalN(t')\neq\emptyset$. If that holds, it means that the robot can reach its goal configuration set $\ccalP_g$. The tree nodes that satisfy this property are collected in a set $\ccalX_g\subseteq\ccalV$ [line \ref{alg1:updXg}, Alg. \ref{alg1}] that is initialized as $\ccalX_g=\emptyset$ [line \ref{alg1:ini_tree}, Alg. \ref{alg1}]. Once the algorithm terminates, we select a tree node $\bbq_g\in\ccalX_g$ (if it is non-empty) [lines \ref{alg1:nonemptySol}-\ref{alg1:pick-qg}, Alg. \ref{alg1}].
Then, we compute the path  $\tau_{\bbq}=\tau_{\bbq}(0),\dots, \tau_{\bbq}(H+1)$ over $\ccalG$ connecting $\bbq_g$ to the tree root $\bbq(0)$, i.e., $\tau_{\bbq}(0)=\bbq(0)$, $\tau_{\bbq}(H+1)=\bbq_g$, and $\tau_{\bbq}(k)\in\ccalV$, $\forall k\in\{1,\dots,H+1\}$ [line \ref{alg1:tauq}, Alg. \ref{alg1}]. Using $\tau_{\bbq}$ and the function $A$, we can extract the plan $\tau=\tau(0),\dots,\tau(H)$ 
[line \ref{alg1:output}, Alg. \ref{alg1}]. The node $\bbq_g$ can be selected randomly or based on any  criterion (e.g., the horizon $H$ of $\tau$). 

\textcolor{black}{Once $\tau = \tau(0), \dots,\tau(k),\dots \tau(H)$ is computed, a robot motion path $\xi$, defined as a sequence of robot configurations $\bbp$, can be constructed sequentially across $k = 0,\dots,H$. Each action $\tau(k) = a(\bbc_i, \bbc_i')$ involves moving an object $o_i$ from its current configuration $\bbc_i$ to a new configuration $\bbc_i'$. Execution of $\tau(k)$ follows the corresponding path $\tau_{\bbq}$: the robot moves towards $o_i$ to grasp it while in $\ccalN$, determined by $\tau_{\bbq}(k) = [\ccalN, \bbc]$, and then drops it at $\bbc_i'$ while in $\ccalN'$, determined by $\tau_{\bbq}(k+1) = [\ccalN', \bbc']$. This path can be computed using any existing motion planner (e.g., RRT). A motion path executing $\tau(k)$ exists because (i) the robot is holonomic and (ii) by construction of Algorithm 2 (i.e., existence of collision-free paths $\zeta$ for each $\tau(k)$). Concatenating these motion paths for all $k$ yields the full robot motion path $\xi$ [line \ref{alg1:motion}, Alg. \ref{alg1}].}

\subsection{LLM-guided Sampling Strategy} \label{M2}

The proposed sampling-based planner relies on the distributions $f_{\ccalV}$, $f_{\ccalO_M}$, and $f_{\ccalC_i}$ to generate $\bbq_{\text{new}}$. While these can be uniform, as in \cite{van2010path}, such strategies often result in slow plan synthesis, particularly as environmental complexity (e.g., the number of movable objects) increases. To accelerate planning, we introduce non-uniform mass functions $f_{\ccalV}$ and $f_{\ccalO_M}$ that bias sampling toward directions where feasible solutions are more likely to be found, while retaining a uniform density for $f_{\ccalC_i}$. To realize this bias, our approach leverages LLMs to identify promising directions. 

First, we define $f_{\ccalV}$ as follows:
\begin{equation}\label{eq:fv}
    f_{\ccalV}(\bbq \mid \ccalV, \ccalV_v)=
    \begin{cases}
        \frac{p_\text{rand}}{|\ccalV_v|} + \frac{1-p_\text{rand}}{|\ccalV|}, & \text{if}~\bbq\in\ccalV_v  \\
        \frac{1-p_\text{rand}}{|\ccalV|}, & \text{otherwise}
    \end{cases}
\end{equation}
where $p_\text{rand}\in[0,1)$ is a user-specified parameter and  $\ccalV_v\subseteq\ccalV$; $\ccalV_v$ will be defined later once $f_{\ccalO_M}$ is defined. 
Notice that if $p_\text{rand}=0$, then $f_{\ccalV}$ models uniform sampling while as $p_{\text{rand}}$ increases, the sampling strategy becomes more biased toward selecting nodes  $\bbq\in\ccalV_v$.

Second, we define the function $f_{\ccalO_M}$ as follows:
\begin{equation}\label{eq:fO}
    f_{\ccalO_M}(o_i \mid o_{\text{LLM}}, \mathcal{O}_{M,\text{rand}})=
    \begin{cases}
        p_\text{obs} + \textcolor{black}{\frac{1-p_\text{obs}}{|\mathcal{O}_{M,\text{rand}}|}}, & \text{if}~o_i=o_{\text{LLM}},  \\
        \frac{1-p_\text{obs}}{|\mathcal{O}_{M,\text{rand}}|}, & \text{if}~o_i\in\mathcal{O}_{M,\text{rand}},
    \end{cases}
\end{equation}
where, $p_\text{obs}\in[0,1)$ is a user-specified parameter, and $o_{\text{LLM}}\in\ccalO_M$ is a manipulable obstacle recommended by an LLM for relocation. How to prompt the LLM to select $o_{\text{LLM}}$ is discussed in Section \ref{M3}. According to \eqref{eq:fO}, the probability of selecting $o_i\notin\ccalO_M$ is $0$. If $p_\text{obs}=0$, $f_{\ccalO_M}$ reduces to uniform sampling; as $p_\text{obs}$ increases, sampling increasingly favors $o_{\text{LLM}}$; \textcolor{black}{see Fig. \ref{fig:distribution}}. If the LLM fails to select a manipulable obstacle or outputs invalid information, we set $p_{\text{obs}} = 0$ for that iteration so a manipulable obstacle is chosen randomly; see Section \ref{M3}. 

Next, we define the set $\ccalV_v$ used to  construct  $f_{\ccalV}$ in \eqref{eq:fv}. First, we define a function $v:\ccalV\rightarrow \mathbb{N}$ assigning a positive score to each tree node $\bbq$. This function is initialized as $v(\bbq(0))=0$. Second, we show how the score of a newly added node $\bbq_{\text{new}}$ is generated; \textcolor{black}{see Fig. \ref{fig:tree_structure}.}
If $\bbq_{\text{new}}$ was constructed by relocating $o_{\text{LLM}}$, 
then we set $v(\bbq_{\text{new}})=v(\bbq_{\text{rand}})+1$. Otherwise, we set $v(\bbq_{\text{new}})=v(\bbq_{\text{rand}})$. Informally, $v(\bbq_{\text{new}})$ counts how many nodes along the path connecting $\bbq_{\text{new}}$ to $\bbq(0)$ have been constructed based on LLM recommendations. Third, we construct $\ccalV_v$ so that it collects the nodes with the highest score, i.e., $\ccalV_v=\{\bbq\in\ccalV~|~v(\bbq)=\max_{\bbq'\in\ccalV} v(\bbq')\}$; \textcolor{black}{see Fig. \ref{fig:tree_structure}.}

\subsection{Prompt Structure} \label{M3}
\begin{figure}
    \centering
    \includegraphics[width=1\linewidth]{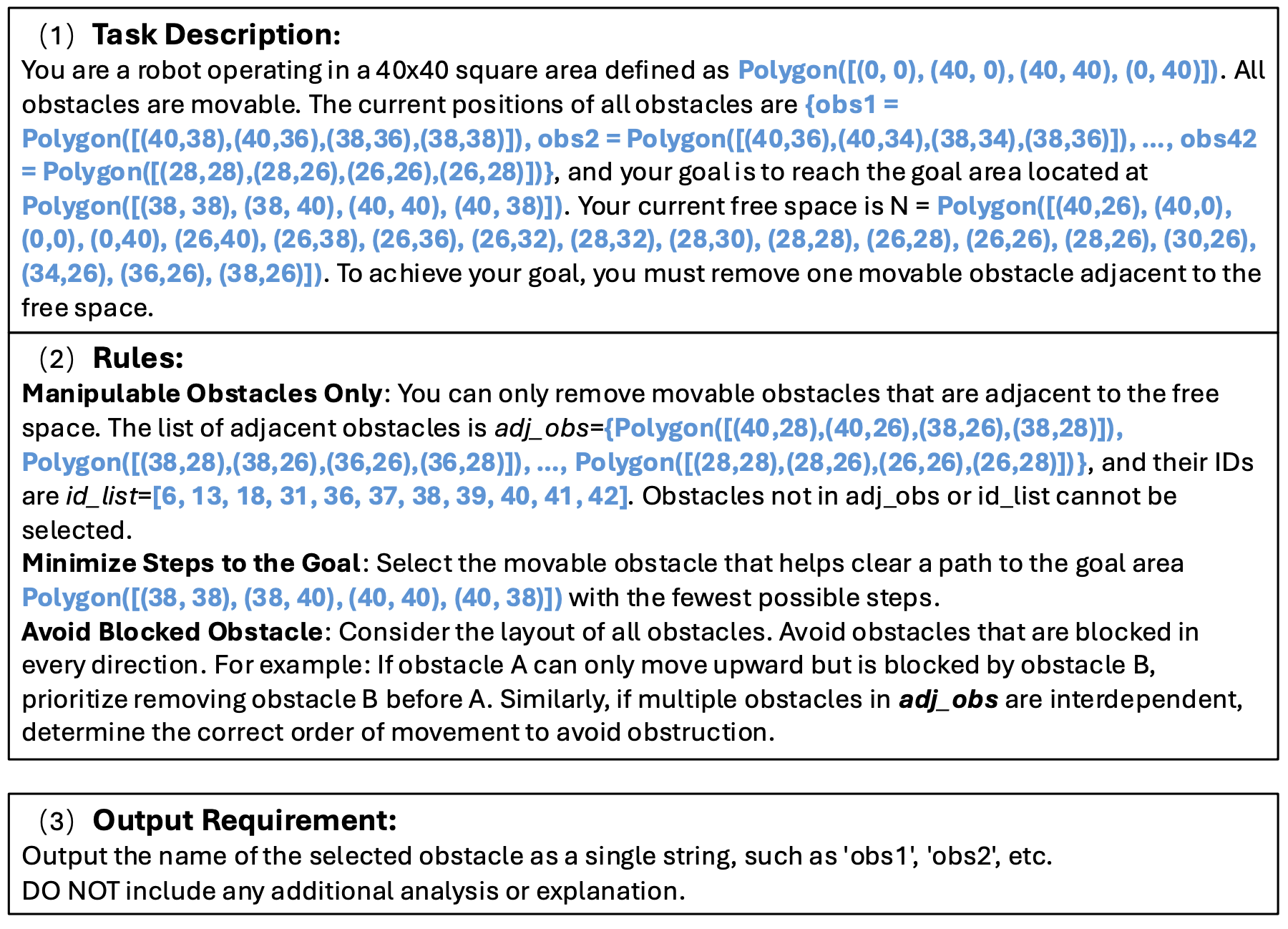}
    \caption{Input prompt for GPT-4o and $K=1$ in Case Study V  (Sec. \ref{sec:sims}). }
    \label{fig:prompt}
\end{figure}

In this section, \textcolor{black}{we describe how we prompt a pre-trained LLM to select a manipulable obstacle $o_{\text{LLM}}$ in a single query.} The input prompt has three parts (see Fig.\ref{fig:prompt}): (1) \textit{Task description}, which specifies the environment (locations of movable/non-movable obstacles and the goal region) and the robot’s task of removing obstacles to reach the goal; (2) \textit{Rules}, which state that the robot may only remove manipulable obstacles, should minimize removals, and should avoid selecting blocked manipulable obstacles (see Case StudyVI in Section~\ref{sec:sims}); and (3) \textit{Output format}, which defines how the LLM’s response must be structured for compatibility with Alg.\ref{alg2}. \textcolor{black}{In particular, the LLM is asked to generate $K \geq 1$ obstacles (in a single query) in the form $\text{obs}i$, where $i$ uniquely identifies an obstacle.} One of the generated obstacles is then selected at random and denoted $o_{\text{LLM}}$ in \eqref{eq:fO}. If $o_{\text{LLM}}$ is not manipulable (content error) or the response does not follow the required structure (format error), we discard the LLM output and set $p_{\text{obs}}=0$ for that iteration.

\subsection{Probabilistic Completeness}\label{sec:probCompl}
Next, we show that NAMO-LLM is probabilistically complete. 
To formally state this result, we first introduce the notion of path clearance \cite{van2010path}. Consider any path $\tau_{\bbq} = \tau_{\bbq}(0), \dots,\tau_{\bbq}(k),\dots, \tau_{\bbq}(H+1)$ computed by Alg. \ref{alg1} yielding a plan $\tau$ solving Problem \ref{pr1}.
We say that $\tau_{\bbq}$ has clearance $\epsilon > 0$ if any alternative path $\tilde{\tau}_{\bbq} = \tilde{\tau}_{\bbq}(0), \dots, \tilde{\tau}_{\bbq}(k),\dots,\tilde{\tau}_{\bbq}(H+1)$ satisfying the following conditions (i)-(ii) for all $k \in \{1, \dots, H\}$ also corresponds to a feasible plan $\tilde{\tau}$. 
(a) The transitions from $\tau_{\bbq}(k-1)$ to $\tau_{\bbq}(k)$ and from $\tilde{\tau}_{\bbq}(k-1)$ to $\tilde{\tau}_{\bbq}(k)$ must involve relocating the same manipulable obstacle $o_i$, albeit potentially to different configurations. (b) Let $\bbc_i^k$ and $\tilde{\bbc}_i^k$ denote the configurations of that obstacle $o_i$ in $\tau_{\bbq}(k)$ and $\tilde{\tau}_{\bbq}(k)$, respectively. These configurations must satisfy $\lVert \bbc_i^k - \tilde{\bbc}_i^k \rVert \leq \epsilon$ for all $k$, for some $\epsilon>0$. In what follows, we refer to satisfaction of conditions (i)–(ii) concisely as $\lVert \tau_{\bbq}(k) - \tilde{\tau}_{\bbq}(k) \rVert \leq \epsilon$. 

\begin{theorem}\label{thm:completeness}
Assume that there exists a feasible path $\tau_{\bbq} =\tau_{\bbq}(0), ...,\tau_{\bbq}(H+1)$ with clearance $\epsilon > 0$. If $p_{\text{rand}}\in[0,1)$, $p_{\text{obs}}\in[0,1)$,  and the motion planner used to check existence of collision-free paths in [line \ref{alg2:check}, Alg. \ref{alg2}] is complete, then the probability that NAMO-LLM will find a feasible path approaches $1$ as $m$ goes to $\infty$.
\end{theorem}

\textit{Proof:}
    To show this result, we need first to introduce the following notations and definitions. Let $\ccalG_m = \{\ccalV_m, \ccalE_m\}$ denote the tree constructed by Algorithm \ref{alg1} at the end of iteration $m$. Since the plan $\tau_{\bbq}$ has clearance $\epsilon > 0$, there exist alternative feasible plans $\tilde{\tau}_{\bbq}$ such that $\lVert \tau_{\bbq}(k) - \tilde{\tau}_{\bbq}(k) \rVert \leq \epsilon$ for all $k$.  Throughout the proof, any reference to a state $\tilde{\tau}_{\bbq}(k)$ assumes this $\epsilon$-closeness to the plan $\tau_{\bbq}$. 

We also define the following two random variables that will be used to model the sampling process and the computation of a feasible plan. First, we define Bernoulli random variables $X_m$ that are equal to $1$ if the following holds for any state $\tilde{\tau}_{\bbq}(k)$ and $k\in \{0,1,\dots,H+1\}$ at iteration $m$:
(i) a state $\tilde{\tau}_{\bbq}(k)$ is added to the tree (i.e., $\tilde{\tau}_{\bbq}(k) \in \ccalV_m$) via an edge from a predecessor $\tilde{\tau}_{\bbq}(k-1) \in \ccalV_{m'}$ for some $m' < m$ (i.e., $(\tilde{\tau}_{\bbq}(k-1), \tilde{\tau}_{\bbq}(k)) \in \ccalE_m$), and
(ii) no state $\tilde{\tau}_{\bbq}'(k)$ satisfying $\lVert \tau_{\bbq}(k) - \tilde{\tau}_{\bbq}'(k) \rVert \leq \epsilon$ was already present in the tree prior to iteration $m$ (i.e., $\tilde{\tau}_{\bbq}'(k) \notin \ccalV_{m'}$ for all $m' < m$).
Additionally, once a state $\tilde{\tau}_{\bbq}(H+1)$ satisfying both (i) and (ii) is added at an iteration $m$, no future iterations can satisfy condition (ii) for any $k$. In this case, by convention, we set $X_{m'} = 1$, for all $m' > m$ (i.e., the probability of success of $X_{m'}$ is $1$). 

Second, we define the random variable $Y_m$ counting the number of successes of the Bernoulli variables $X_1, X_2,\dots,X_m$, i.e., $Y_m=\sum_{i=1}^m X_i$. By construction of $X_i$, if $Y_m > H$, then a state 
$\tilde{\tau}_{\bbq}(H+1)$ has been added to $\ccalV_m$ indicating that a solution has been found. Therefore, to show probabilistic completeness, it suffices to show that the probability $\mathbb{P}(Y_m \geq H+1)$ approaches $1$ as $m \to \infty$.

Using the Chebyshev inequality, we can bound the “failure” probability, i.e., the probability that fewer than $H$ successful expansions occurred in $m$ iterations (which means the goal has not been reached), as follows:

\begin{align}\label{eq:chebYm}
\mathbb{P}(Y_m \leq H) & \leq \mathbb{P}\Bigl( |Y_m - \mathrm{E}(Y_m)| \geq \mathrm{E}(Y_m) - H \Bigl) \nonumber\\
& \leq \frac{\mathrm{Var}(Y_m)}{(\mathrm{E}(Y_m) - H)^2}
\end{align}

The expectation  $\mathrm{E}(Y_m)$ and variance $\mathrm{Var}(Y_m)$ of $Y_m$ are defined as follows
\begin{equation}\label{eq:exYm}
    \mathrm{E}(Y_m) = \sum_{i=1}^{m}\mathrm{E}(X_i) = \sum_{i=1}^{m}p_i
\end{equation}
and 
\begin{align}\label{eq:VarYm}
&\mathrm{Var}(Y_m) = \sum_{i=1}^{m}\mathrm{Var}(X_i) + 2\sum_{1\leq i<j \leq m}\mathrm{Cov}(X_i, X_j) \nonumber\\
&= \sum_{i=1}^{m}p_i(1-p_i) + 2\sum_{1\leq i<j \leq m}\Bigl[\mathrm{E}(X_iX_j)-p_ip_j\Bigr],
\end{align}
where $p_i$ is the probability of success of $X_i$.

In what follows, we show (a) $\lim_{m\to\infty} E(Y_m)=\infty$ and (b) $\mathrm{Var}(Y_m)<3\mathrm{E}(Y_m)$. Using (b), we can re-write \eqref{eq:chebYm} as follows:
\begin{equation}\label{eq:chebYm2}
    \mathbb{P}(Y_m \leq H) \leq \frac{\mathrm{Var}(Y_m)}{(\mathrm{E}(Y_m) - H)^2}\leq \frac{3\mathrm{E}(Y_m)}{(\mathrm{E}(Y_m) - H)^2}.
\end{equation}
Then combining (a) with \eqref{eq:chebYm2}, we get that $\lim_{m\to\infty} \mathbb{P}(Y_m \leq H)=0$. This equivalently means $\lim_{m \to \infty} \mathbb{P}(Y_m \geq H+1) = 1$ completing the proof.

\textbf{Proof of (a):} Here we show that $\lim_{m\to\infty} E(Y_m)=\infty$.
\underline{Step 1:} To show this, we first show the probability of success of $X_m$, denoted by $p_m$, is strictly positive, i.e., $p_m>0$, for all $m$. This result will later be used to establish a lower bound on $\mathbb{P}(Y_m \geq H+1)$. Showing this consists of the following two parts.

\textit{Part 1:} First, let $\hat{\ccalV}_m\subseteq \ccalV_m$ collect all states $\tilde{\tau}_{\bbq}(k)\in\ccalV_m$ for which there does not exist a state $\tilde{\tau}_{\bbq}(k')$ with  $k'>k$ that is already in the set $\ccalV_m$. A necessary requirement for $X_m=1$ is to select a state from $\hat{\ccalV}_{m-1}$ to be $\bbq_{\text{rand}}$. By construction of $f_{\ccalV}(\bbq \mid \ccalV, \ccalV_v)$, the probability $p_1^m=f_{\ccalV}(\bbq \mid \ccalV, \ccalV_v)$ of selecting a state $\bbq_{\text{rand}}\in \hat{\ccalV}_{m-1}$ is strictly positive. Specifically, we have that $p_1^m \geq \min\{\frac{p_\text{rand}}{|\ccalV_v|} + \frac{1-p_\text{rand}}{|\ccalV_{m-1}|},\frac{1-p_\text{rand}}{|\ccalV_{m-1}|} \} = \frac{1-p_\text{rand}}{|\ccalV_{m-1}|}\geq \frac{1-p_\text{rand}}{m}$ where the last inequality is due to $|\ccalV_{m-1}|\leq {m}$. Observe that $p_2^m>0$ since by assumption $p_\text{rand}<1$. 

\textit{Part 2:} Second, assume that a node $\tilde{\tau}_{\bbq}(k)$ from $\hat{\ccalV}_m$ has been selected to be $\bbq_{\text{rand}}$. Under this assumption, there exist nodes $\bbq_{\text{new}}$ that can be added to the tree satisfying conditions (i)-(ii) required for $X_m=1$. Specifically, all nodes $\tilde{\tau}_{\bbq}(k+1)$ are such candidate nodes. Next, we show that the probability $p_2^m$ of sampling a state $\tilde{\tau}_{\bbq}(k+1)$ to be $q_{\text{new}}$ is strictly positive, i.e., $p_2^m>0$, for all $m$. Given $\bbq_{\text{rand}}=\tilde{\tau}_{\bbq}(k)$, sampling a state $\bbq_{\text{new}}=\tilde{\tau}_{\bbq}(k+1)$ will occur under the following two requirements (as per the $\epsilon$-clearance definition). First, at state $\bbq_{\text{rand}}=\tilde{\tau}_{\bbq}(k)$ the obstacle $o_i$ that will be selected to relocate is the one selected in the transition from $\tau_{\bbq}(k)$ to $\tau_{\bbq}(k+1)$. By construction of $f_{\ccalO_M}(o_i \mid o_{\text{LLM}}, \mathcal{O}_{M,\text{rand}})$, since the obstacle $o_i$ is manipulable (by construction of $\tau_{\bbq}$) this will occur with probability $p_{o_i}^m\geq \min\{ p_\text{obs} + \frac{1}{|\mathcal{O}_{M,\text{rand}}|},\frac{1-p_\text{obs}}{|\mathcal{O}_{M,\text{rand}}|}\}$. Observe that since $p_{\text{obs}}<1$, we have that $p_{o_i}^m>0$ for all $m$. Second, assuming that $o_i$ will be selected for relocation, its new configuration, denoted by $\tilde{c}_i^{k+1}$, should satisfy $\lVert \tilde{c}_i^{k+1}-c_i^{k+1} \rVert \textcolor{black}{\leq \epsilon} $ where $c_i^{k+1}$ is the configuration of $o_i$ at state $\tau_{\bbq}(k+1)$. Since by assumption $\epsilon>0$ and $f_{\ccalC_i}$ is a uniform density function, we get that the probability $p_{c_i}^m$ of sampling a $\tilde{c}_i^{k+1}$ that satisfies this requirement is also strictly positive for all $m$. Thus, we get that $p_2^m=p_{o_i}^m p_{c_i}^m>0$. Consequently, there must exist a positive constant $p_2^\text{min}$ such that $p_2^m \geq p_2^\text{min}$ for all $m\geq 0$ 
We note also since a complete planner is used to check if there exists a collision-free path reaching $\tilde{c}_i^{k+1}$ from $\tilde{c}_i^{k}$ (i.e., the configuration of the obstacle $o_i$ in $\tilde{\tau}_{\bbq}(k)$), then the sampled configuration $\tilde{c}_i^{k+1}$ will be never be rejected in line \ref{alg2:check} incorrectly (i.e., if such a collision-free path exists). 

Combining Part 1 and 2, we get that the probability $p_m$ of success of $X_m$ is strictly positive since $p_m=p_1^mp_2^m\geq \frac{(1-p_\text{rand})p_2^m}{m} \geq \frac{(1-p_\text{rand})p_2^\text{min}}{m} >0.$ 


\underline{Step 2:} Given the result from Step 1, we can re-write the expected value of $Y_m$ in \eqref{eq:exYm} as:

\[E(Y_m) = \sum_{i=1}^{m}p_1^ip_2^i\geq \sum_{i=1}^{m}\frac{(1-p_\text{rand})p_2^\text{min}}{i} =\alpha \sum_{i=1}^{m}\frac{1}{i}\]
where \textcolor{black}{$\alpha=(1-p_\text{rand})p_2^\text{min}>0$ is a positive constant across all iterations $m$}. Then, 
\[\mathrm{E}(Y_m) \geq \alpha \sum_{i=1}^{m}\frac{1}{i}\]
When $m \to \infty$, $\sum_{i=1}^{m}\frac{1}{i} \to \infty$. Therefore, we get $\mathrm{E}(Y_m) \to \infty$ completing the proof of part (a).

\textbf{Proof of (b):} Here we show that $\mathrm{Var}(Y_m)<3\mathrm{E}(Y_m)$. Recall from \eqref{eq:VarYm} that the variance of $Y_m$ is given by:

\begin{align}\label{eq:Var1}
\mathrm{Var}(Y_m) &=\sum_{i=1}^{m}p_i(1-p_i) + 2\sum_{1\leq i<j \leq m}\Bigl[E(X_iX_j)-p_ip_j\Bigr]
\end{align}

First, we focus on the term $\sum_{i=1}^{m}p_i(1-p_i)$ in \eqref{eq:Var1}. Since $0 < p_i \leq 1$, we have that 
\[\sum_{i=1}^{m}p_i(1-p_i) < \sum_{i=1}^{m}p_i = \mathrm{E}(Y_m) \]

Second, we focus on the term $\sum_{1\leq i<j \leq m}\Bigl[E(X_iX_j)-p_ip_j\Bigr]$ in \eqref{eq:Var1}. Since $X_iX_j \leq X_i$, we have that:
\[\mathrm{E}(X_iX_j) \leq \mathrm{E}(X_i)=p_i\]
Hence,
\[
\mathrm{E}(X_iX_j) -p_ip_j \leq p_i(1-p_j) < p_i
\]
Then, \[
2\sum_{1\leq i<j \leq m}\Bigl[\mathrm{E}(X_iX_j)-p_ip_j\Bigr] < 2\sum_{i=1}^{m}p_i  = 2 \mathrm{E}(Y_k)
\]
Thus we get $\mathrm{Var}(Y_m) < 3\mathrm{E}(Y_m)$ completing the proof of part (b).

\vspace{-0.1cm}
\section{Experiments}\label{sec:sims}
\vspace{-0.1cm}

\textcolor{black}{In this section, we demonstrate that NAMO-LLM outperforms search-based methods \cite{ellis2023navigation}, sampling-based planners \cite{van2010path}, and LLM-based planners in terms of runtimes, plan quality, and task success rates. Our evaluation is performed across different LLMs and environmental setups.
Hardware demonstrations of our method on ground robots are also provided; see also the supplemental material.} 
All experiments were conducted on a computer with an Intel Core i5 2.4GHz processor and 8 GB of RAM. Software implementation of NAMO-LLM can be found in \cite{codeNAMOLLM}.

\subsection{Setting Up Comparative Experiments}\label{sec:setup}
\textbf{NAMO Baselines:} \textcolor{black}{We consider three baselines: (i) RandomTree, a sampling-based approach from \cite{van2010path} that employs uniform sampling; (ii) NAMO-SA, a search-based method from \cite{ellis2023navigation}; and (iii) an LLM-based planner that relies exclusively on LLMs. Among these, only (iii) incorporates LLMs. Comparisons with (i)–(ii) are reported in Sections \ref{sec:comp}–\ref{sec:variousLLMs}, and comparisons with (iii) are presented in Section \ref{sec:compLLM-Planner}.}

\textcolor{black}{RandomTree is recovered from our method by setting $p_{\text{rand}}=p_{\text{obs}}=0$, while the LLM-based planner corresponds to NAMO-LLM with $p_{\text{rand}}=p_{\text{obs}}=1$ and $K=1$. Unlike our method and RandomTree, the LLM-based planner lacks completeness guarantees (see Theorem~\ref{thm:completeness}).
RandomTree and NAMO-LLM run until either a feasible solution is found or a maximum of $m_{\text{max}}=30,000$ iterations is reached. In the LLM-based planner, the LLM is queried at every iteration to select an obstacle, which is placed at a random location. 
To ensure financially sustainable but fair comparisons, we cap the total number of LLM queries to $3H_\text{min}$, where $H_\text{min}$ is the minimum number of obstacles that need to be relocated to clear a path to the goal area.
At each iteration, the LLM receives the same prompt as NAMO-LLM and outputs an obstacle. If the output has no format/content errors, the obstacle is placed randomly; otherwise, the counter is incremented and the LLM is queried again.}

The approach in \cite{ellis2023navigation} selects obstacles for removal by computing the total distance from the robot to the obstacle and from the obstacle to the goal region using $\text{A}^*$. The obstacle with the lowest cost is chosen and moved to a \textit{pre-defined storage room} if a collision-free path exists; otherwise, the next lowest-cost obstacle is considered. In our implementation, the storage room is modeled as an external region adjacent to the workspace boundary, and once an obstacle reaches this boundary it is assumed to be instantaneously transferred. 
In contrast, our method designs custom drop-off locations during sampling. This favors the baseline, since our method may sample invalid drop-off locations
leading to rejections; see lines \ref{alg2:check0}–\ref{alg2:check} in Alg.~\ref{alg2}. Moreover, in \cite{ellis2023navigation}, once an obstacle is moved to the storage room it is never relocated, further favoring the baseline as the number $n$ of movable obstacles decreases over time—unlike in our method. Thus, the maximum horizon $H$ of a plan in \cite{ellis2023navigation} equals $n$, whereas in ours it is possible that $H>n$ since obstacles may be reconfigured multiple times. Re-planning strategies  in \cite{ellis2023navigation} to handle unknown obstacles are neglected in our implementation as we focus on known environments. 
Finally, to ensure fair comparisons, our implementation of NAMO-LLM ([line \ref{alg2:check}, Alg. \ref{alg2}]), \cite{ellis2023navigation}, and \cite{van2010path} share the same RRT planner to check if there exists a collision-free path towards the drop-off location of an obstacle and the storage area.

\textbf{Robot and Obstacle Configuration Space:} We consider a disk-shaped robot of radius $r$ ($r=0.2$ in Cases I–IV, $r=0.5$ in Cases V–VI) with configuration space $\ccalP$ given by the robot’s center position. Obstacles are polygonal, each with configuration space $\ccalC_i$ defined by its position and orientation.

\textbf{Evaluation Metrics:} \textcolor{black}{As evaluation metrics we use (a) the runtime to compute the first feasible plan; (b) the horizon $H$ of the designed plan; and (c) task success rates. Task success rates, runtimes, and plan horizons $H$ for our method, the LLM-based planner, and \cite{van2010path} are averaged over $100$ runs. Task success rates are defined as the percentage of runs that result in a feasible manipulation plan. Note that NAMO-SA’s success rate is always $100\%$, since it has no randomized component and, in the worst case, moves all objects sequentially to a storage room as discussed earlier. RandomTree and NAMO-LLM will also have a $100\%$ success rate, for large enough $m_{\text{max}}$, due to their completeness properties. Our reported times include all computational components of Alg. 1–2 including e.g., the runtime for LLM inference. }

\begin{figure}[t]
\centering
\includegraphics[width=0.8\linewidth]{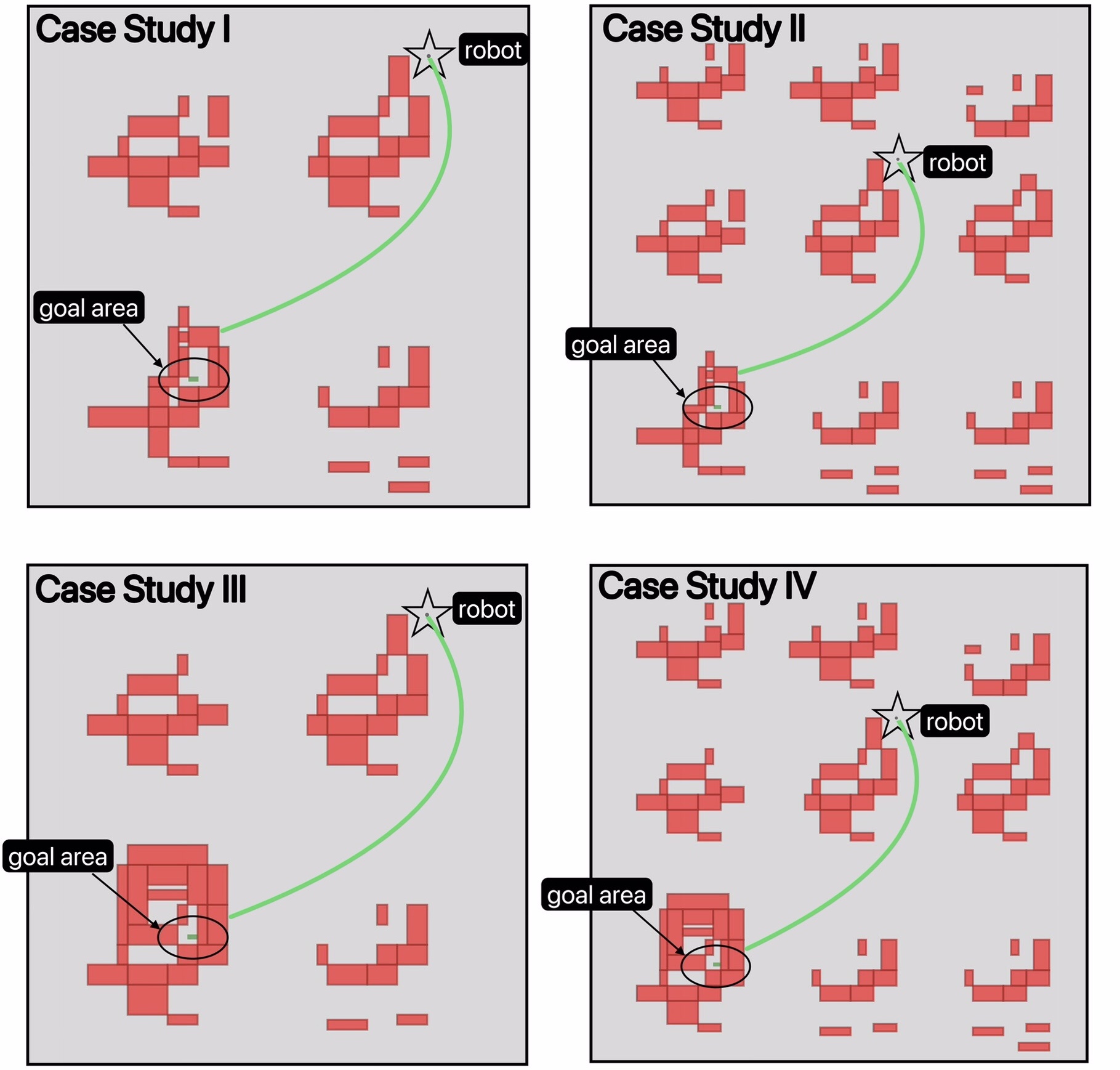} 
\vspace{-0.2cm}
\caption{Initial configurations of the obstacles (red polygons) and robot for Case Studies \Romannum{1}-\Romannum{4}. The green path connects the initial robot state to the first obstacle to be repositioned for clearing a path to the goal area $\ccalP_g$.}
\vspace{-0.2cm}
\label{case1_4}
\end{figure}

\subsection{\textcolor{black}{Comparisons with Search- and Sampling-Based Baselines}}\label{sec:comp}

\textcolor{black}{We present six case studies that vary in complexity based on the total number $n$ of movable obstacles, their initial configurations, and the minimum plan horizon $H_{\text{min}}$ required to reach $\ccalP_g$. For brevity, we refer to our method as NAMO-LLM$(p_\text{rand}, p_\text{obs})$. In what follows, we pair NAMO-LLM$(p_\text{rand}, p_\text{obs})$ with the pre-trained GPT-4o model.} \textcolor{black}{All methods achieve $100\%$ task success rates. The only exception was RandomTree achieving $0\%$ in Case V.}

\begin{table}[t]
\centering
\begin{tabular}{|c|c|c|c|c|}
\hline
& Method & $|\ccalV|$ & Time (secs) & $H$ \\ \noalign{\hrule height 1pt}
& NAMO-SA \cite{ellis2023navigation} & N/A & 99.05 & 27 \\ \hline
& RandomTree \cite{van2010path} & 15.3 & 14.94 & 4.3 \\ \noalign{\hrule height 1pt}
\multirow{4}{*}{K=1} 
& NAMO-LLM(0.2, 0.2) & 6.6 & 6.42 & 3.1 \\ \cline{2-5}
& NAMO-LLM(0.2, 0.8) & 3.15 & 3.21 & \textbf{2.5} \\ \cline{2-5}
& NAMO-LLM(0.8, 0.2) & 6.26 & 5.74 & 3.74 \\ \cline{2-5}
& NAMO-LLM(0.8, 0.8) & 2.71 & \textbf{2.88} & 2.52 \\ \noalign{\hrule height 1pt}
\multirow{4}{*}{K=2} 
& NAMO-LLM(0.2, 0.2) & 7.85 & 8.77 & 3.6 \\ \cline{2-5}
& NAMO-LLM(0.2, 0.8) & 3.30 & 4.91 & \textbf{2.65} \\ \cline{2-5}
& NAMO-LLM(0.8, 0.2) & 5.7 & 4.94 & 3.5 \\ \cline{2-5}
& NAMO-LLM(0.8, 0.8) & 3.11 & \textbf{4.46} & 2.89 \\ \noalign{\hrule height 1pt}
\multirow{4}{*}{K=3} 
& NAMO-LLM(0.2, 0.2) & 7.05 & 7.68 & 3.10 \\ \cline{2-5}
& NAMO-LLM(0.2, 0.8) & 3.30 & 4.47 & \textbf{2.43} \\ \cline{2-5}
& NAMO-LLM(0.8, 0.2) & 5.45 & 5.14 & 3.35 \\ \cline{2-5}
& NAMO-LLM(0.8, 0.8) & 3.21 & \textbf{4.25} & 2.67 \\ \hline
\end{tabular}
\caption{Case Study \Romannum{1}: Average tree size $|\ccalV|$, runtimes, and horizon $H$.}
\label{tab:tabI}
\vspace{-0.5cm}
\end{table}

\begin{table}[t]
\centering
\vspace{0.2cm}
\begin{tabular}{|c|c|c|c|c|}
\hline
& Method & $|\ccalV|$ & Time (secs) & $H$ \\ \noalign{\hrule height 1pt}
\multirow{6}{*}{Case \Romannum{2}} 
& NAMO-SA \cite{ellis2023navigation} & N/A & 876.36 & 65 \\ \cline{2-5}
& RandomTree \cite{van2010path} & 34.90 & 42.78 & 4.57 \\ \Xcline{2-5}{0.8pt}
& NAMO-LLM(0.2, 0.8) & 3.86 & 9.99 & \textbf{2.75} \\ \cline{2-5}
& NAMO-LLM(0.8, 0.8) & 4.47 & \textbf{7.80} & 3.04 \\ \noalign{\hrule height 1pt}
\multirow{6}{*}{Case \Romannum{3}} 
& NAMO-SA \cite{ellis2023navigation} & N/A & 151.23 & 26 \\ \cline{2-5}
& RandomTree \cite{van2010path} & 92.72 & 71.32 & 5.72 \\ \Xcline{2-5}{0.8pt}
& NAMO-LLM(0.2, 0.8) & 6.33 & 8.52 & \textbf{3.70} \\ \cline{2-5}
& NAMO-LLM(0.8, 0.8) & 4.37 & \textbf{5.91} & 3.74 \\ \noalign{\hrule height 1pt}
\multirow{6}{*}{Case \Romannum{4}} 
& NAMO-SA \cite{ellis2023navigation} & N/A & 877.97 & 65 \\ \cline{2-5}
& RandomTree \cite{van2010path} & 290.83 & 408.68 & 7.41 \\ \Xcline{2-5}{0.8pt}
& NAMO-LLM(0.2, 0.8) & 8.23 & 19.31 & \textbf{4.16} \\ \cline{2-5}
& NAMO-LLM(0.8, 0.8) & 5.70 & \textbf{17.12} & 4.5 \\ \hline
\end{tabular}
\caption{Case Study \Romannum{2}-\Romannum{4}: Average tree size, runtimes, and horizon.}
\label{tab:tabII}
\vspace{-0.5cm}
\end{table}

\textbf{Case Study I} ($n=50, H_{\text{min}}=1$): We consider a $50 \times 50$ environment with $n = 50$ movable obstacles admitting a feasible plan with  $H_{\text{min}} = 1$; see Fig. \ref{case1_4}.
We compared NAMO-LLM for various values of $p_{\text{rand}}$, $p_{\text{obs}}$, and $K$ against the baselines; see Table \ref{tab:tabI}. Observe that our method significantly outperforms both baselines in terms of both the runtime required to compute the first feasible plan and the horizon $H$ of that plan. 
For instance, NAMO-LLM tends to be 30x and 5x faster than \cite{van2010path} and \cite{ellis2023navigation}, respectively.
Observe that the performance of our method does not change significantly with respect to $K\in\{1,2,3\}$. We attribute this to the fact that there are multiple candidate obstacles near the goal area $\ccalP_g$ that can be removed to get access to it and that the employed LLM is capable of identifying them. 

In Case Studies II-V, we present results for our method with $K=1$ combined with $p_{\text{rand}}=p_{\text{obs}}=0.8$, and $p_{\text{rand}}=0.2$, $p_{\text{obs}}=0.8$. 


In \textbf{Case Study II} ($n=100, H_{\text{min}}=1$), \textbf{Case Study III} ($n=50, H_{\text{min}}=2$) and \textbf{Case Study IV} ($n=100, H_{\text{min}}=2$), we similarly observe that the performance gap between our method and the baselines further increases as the minimum required horizon $H_{\text{min}}$ grows and/or $n$ increase. The environments associated with these case studies can be found in Fig \ref{case1_4} while the results are shown in Table \ref{tab:tabII}. 



\begin{figure}[t]
\centering
    \includegraphics[width=0.75\linewidth]{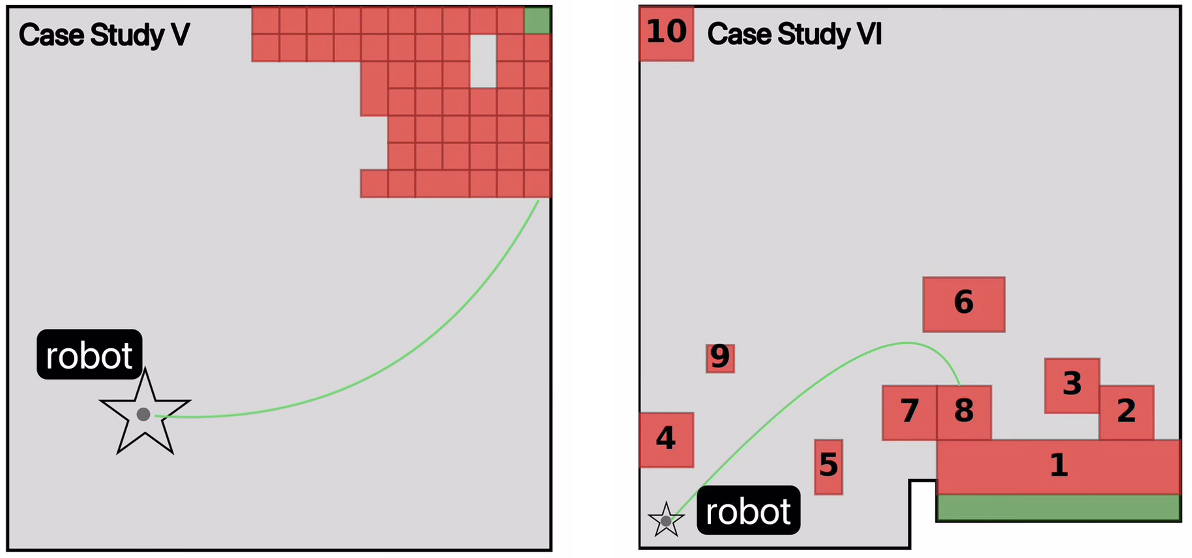} 
    \vspace{-0.25cm}
\caption{Case Study \Romannum{5} \& \Romannum{6}: Initial configurations of the obstacles (red polygons). The green area corresponds to $\ccalP_g$.
}
\label{ex56}
\end{figure}

\textbf{Case Study \Romannum{5}} ($n=50, H_{\text{min}}=6$): We consider a $40 \times 40$ environment; see Fig. \ref{ex56}. 
This scenario is significantly more complex due to the larger number of obstacles that need to be relocated; this is also reflected in the increased runtimes reported in Table \ref{tab:tabV}, compared to the previous case studies. Our method, when configured with $K=1$, $p_{\text{rand}}=p_{\text{obs}}=0.8$ computed the first feasible plan in $48.22$ secs, with an average horizon of $H=26.48$. In contrast, \cite{van2010path} failed to find a feasible plan within $6$ hours while \cite{ellis2023navigation} can identify a feasible plan within $141.91$ secs with $H=42$. 

%



\textbf{Case Study  \Romannum{6}} ($n=10, H_{\text{min}}=3$): 
Finally, we consider a more complex scenario to illustrate the utility of parameter $K$.
Unlike previous case studies, here some manipulable obstacles are blocked by others. 
Specifically, we consider a $20 \times 20$ environment with $n=10$ movable obstacles, where reaching $\mathcal{P}_g$ requires relocating $o_1$, but no collision-free path to any $\bbc_1'$ exists until $o_8$ and $o_2$ are moved; see Fig. \ref{ex56}. Table \ref{tab:tabVI} shows that, unlike Case I, performance improves significantly for $K \in \{2,3\}$ compared to $K=1$. 
This occurs because because with $K=1$ the LLM often prioritizes $o_1$, which is blocked by $o_8$ and $o_2$; any $\bbc_1'$ sampled for $o_1$ is rejected at line \ref{alg2:check} in Alg. \ref{alg2}, slowing progress. However, for $K>1$, the LLM outputs a set of candidate obstacles (often including $o_2$ and/or $o_8$), and random selection from this set helps avoid persistent incorrect recommendations.
Even with $K=1$, our method outperforms \cite{van2010path} in runtime because the LLM occasionally recommends $o_2$ and $o_8$ after some exploratory moves (e.g., relocating $o_3$), which alters its prompt.
Overall, our method surpasses both baselines for $K=2,3$; note in Table \ref{tab:tabVI} that \cite{ellis2023navigation} moved all obstacles to storage since $H=10$. 

\textbf{Summary:} The performance gap between NAMO-LLM and the baselines increases as $n$ or $H_{\text{min}}$ grows. 
Our method achieves its lowest runtime for computing the first feasible plan when $p_{\text{rand}}=p_{\text{obs}}=0.8$ and $K=1$, provided manipulable obstacles are not blocked (Cases I–V). This is due to the LLM’s ability to identify the correct obstacle to remove in these scenarios.
When $p_{\text{rand}}=0.2$ and $p_{\text{obs}}=0.8$, plans typically have shorter horizons, since more uniform node expansion leads to higher-quality first plans, albeit with slightly longer runtimes.
We also observed that LLMs may produce infeasible recommendations when manipulable obstacles are blocked (e.g., Case VI). In such cases, setting $K>1$ prevents the algorithm from stalling on poor LLM predictions.
\textcolor{black}{Finally, across all case studies, $2.77\%$ of LLM outputs are invalid (format or content errors) on average, while $\sim70\%$ of which are due to content errors; these are handled via uniform sampling as discussed in Section \ref{M3}.}

\begin{table}[t]
\centering
\begin{tabular}{|c|c|c|c|}
\hline
 Method & $|\ccalV|$ & Time (secs) & $H$ \\ \noalign{\hrule height 1pt}
NAMO-SA \cite{ellis2023navigation} & N/A & 141.91 & 42 \\ \hline
RandomTree \cite{van2010path} & over 30k & over 22k & N/A \\ \noalign{\hrule height 1pt}
NAMO-LLM(0.2, 0.8) & 106.12 & 143.83 & \textbf{19.65} \\ \hline
NAMO-LLM(0.8, 0.8) & 34.78 & \textbf{48.22} & 26.48 \\ \hline
\end{tabular}
\caption{Case Study \Romannum{5}: Average tree size, runtimes, and horizon. 
}
\vspace{-0.7cm}
\label{tab:tabV}
\end{table}

\begin{table}[t]
\centering
\begin{tabular}{|c|c|c|c|c|}
\hline
 & Method & $|\ccalV|$ & Time (secs) & $H$  \\ \noalign{\hrule height 1pt}
& NAMO-SA \cite{ellis2023navigation} & N/A & 87.28 & 10 \\ \hline
& RandomTree \cite{van2010path} & 326.65 & 227.46 & 8.95 \\ \noalign{\hrule height 1pt}
 \multirow{4}{*}{K=1} 
 & NAMO-LLM(0.2, 0.2) & 112.45 & 188.95 & 10.1 \\ \cline{2-5}
 & NAMO-LLM(0.2, 0.8) & 32.8 & 359.60 & \textbf{8.11}  \\ \cline{2-5}
 & NAMO-LLM(0.8, 0.2) & 85.6 & \textbf{162.09} & 10.8 \\ \cline{2-5}
 & NAMO-LLM(0.8, 0.8) & 18.45 & 187.31 & 8.54 \\ \noalign{\hrule height 1pt}
 \multirow{4}{*}{K=2} 
 & NAMO-LLM(0.2, 0.2) & 82.9 & 81.92 & 10.6 \\ \cline{2-5}
 & NAMO-LLM(0.2, 0.8) & 18.5 & 52.66 & \textbf{7.95}  \\ \cline{2-5}
 & NAMO-LLM(0.8, 0.2) & 32.3 & 37.69 & 11.1 \\ \cline{2-5}
 & NAMO-LLM(0.8, 0.8) & 12.7& \textbf{31.05} & 9.5 \\ \noalign{\hrule height 1pt}
 \multirow{4}{*}{K=3}
 & NAMO-LLM(0.2, 0.2) & 70.95 & 98.02 & 10.65 \\ \cline{2-5}
 & NAMO-LLM(0.2, 0.8) & 26.25 & 93.51 & \textbf{8.35}  \\ \cline{2-5}
 & NAMO-LLM(0.8, 0.2) & 34.9 & 44.23 & 12.05 \\ \cline{2-5}
 & NAMO-LLM(0.8, 0.8) & 12.65 & \textbf{42.47} & 9.9 \\ \hline
 \end{tabular}
 \caption{Case Study \Romannum{6}: Average tree size, runtimes, and horizon.}
 \vspace{-0.2cm}
 \label{tab:tabVI}
 \end{table}

\begin{table}[t]
\centering
\begin{tabular}{|c|c|c|c|c|}
\hline
& Method  & $|\ccalV|$ & Time (secs) & $H$ \\ \noalign{\hrule height 1pt}
\multirow{7}{*}{Case \Romannum{4}} 
& NAMO-SA \cite{ellis2023navigation} & N/A & 877.97 & 65 \\ \cline{2-5}
& RandomTree \cite{van2010path} & 290.83 & 408.68 & 7.41 \\ \cline{2-5}
& GPT 4o & 5.70 & 17.12 & 4.16 \\ \cline{2-5}
& MiniGPT4o(0.2, 0.8) & 19.98 & 81.45 & \textbf{5.62} \\ \cline{2-5}
& MiniGPT4o(0.8, 0.8) & 10.18 & \textbf{59.72} & 6.96 \\ \cline{2-5}
& Gemini(0.2, 0.8) & 84.35 & 233.94 & \textbf{15.8} \\ \cline{2-5}
& Gemini(0.8, 0.8) & 24.75 & \textbf{104.30} & 17.2 \\ \noalign{\hrule height 1pt}
\multirow{7}{*}{Case \Romannum{5}}
& NAMO-SA \cite{ellis2023navigation} & N/A & 141.91 & 42 \\ \cline{2-5}
& RandomTree \cite{van2010path} & over 30k & over 22k & N/A \\ \cline{2-5}
& GPT 4o & 34.78 & 48.22 & 19.65 \\ \cline{2-5}
& MiniGPT4o(0.2, 0.8) & 52.1 & 67.61 & \textbf{13.23} \\ \cline{2-5}
& MiniGPT4o(0.8, 0.8) & 20.3 & \textbf{27.13} & 13.57 \\ \cline{2-5}
& Gemini(0.2, 0.8) & 27.70 & 21.03 & \textbf{9.6} \\ \cline{2-5}
& Gemini(0.8, 0.8) & 15.15 & \textbf{14.24} & 12.15 \\ \hline
\end{tabular}
\caption{\textcolor{black}{Performance comparison of different models in Cases \Romannum{4} and \Romannum{5}, showing average tree size, runtimes, and horizon. The line `GPT 4o' reports the best results of our method when paired with GPT 4o. Lines of the form `LLM($p_{\text{rand}},p_{\text{obs}}$)' refer to our method when coupled with the respective LLM.} }
\label{comparison}
\vspace{-0.5cm}
\end{table}

\subsection{Comparative Results with Alternative LLMs \& VLMs} \label{sec:variousLLMs}
\textcolor{black}{We evaluate NAMO-LLM with GPT-4o, GPT-4o-mini, and Gemini-1.5-Flash on Cases IV and V; see the Table \ref{comparison}). Switching among these models affects performance, but our method remains more scalable than the baselines. In Case Study IV, it outperforms both baselines in runtime for all tested LLMs, though with the Gemini model it produces longer plans than \cite{van2010path} (while still shorter than those of \cite{ellis2023navigation}). In Case Study V, our method outperforms both baselines in both metrics for every LLM tested.} 
\textcolor{black}{We also observed that the performance of our method deteriorates when it is paired with VLMs (GPT 4o). For instance, our average runtime and horizon on Case IV with $p_{\text{rand}}=p_{\text{obs}}=0.8$ is $252.21$ secs and $32.8$, respectively, while it fails to address Case V within $1,000$ secs. The VLM is prompted with an overhead image of the environment, replacing the textual description of the environment. We attribute this result to potentially limited VLM capabilities in object recognition and counting tasks.}



\subsection{Comparisons with LLM-based Planners}\label{sec:compLLM-Planner}
\textcolor{black}{We compare NAMO-LLM against the LLM-based planner from Section \ref{sec:setup}, using GPT-4o.
The baseline’s success rates in Cases I–VI are $86\%$, $92\%$, $78\%$, $90\%$, $54\%$, and $0\%$, respectively, while NAMO-LLM achieves $100\%$ in all. GPT-4o alone is competitive in simpler tasks (Cases I–IV) but drops sharply in complex ones (Cases V–VI).}\footnote{\textcolor{black}{Other LLMs showed the same trend; e.g., GPT-4o-mini achieved 81\%, 70\%, 92\%, 82\%, 55\%, and 0\% for Cases \Romannum{1}–\Romannum{6}, respectively. }} 
\textcolor{black}{For Cases I–IV, both methods produce plans with comparable horizons. As complexity increases (Cases V–VI), NAMO-LLM outperforms the baseline. Across Cases I–VI, the best (average) horizons of NAMO-LLM are $2.5$, $2.75$, $3.70$, $4.16$, $19.65$, and $8.11$, compared to $2.68$, $2.7$, $3.4$, $3.75$, $52.88$, and N/A for the baseline. 
For runtimes, the baseline is slightly faster in simpler tasks but fails on more complex ones (Case V): NAMO-LLM achieves $2.88$, $7.80$, $5.91$, $17.12$, and $31.05$ secs across Cases I–V, compared to $2.73$, $6.47$, $5.37$, $13.17$, and N/A for the baseline.}
\textcolor{black}{Overall, these comparisons show that our algorithm consistently outperforms the baseline in terms of task success rates, particularly in complex scenarios (e.g., Case V). The main advantage of the baseline is that, \textit{if} it produces a plan, the plan may be found slightly faster 
especially in simpler tasks (Cases I–V).}

\subsection{\textcolor{black}{Hardware Validation Example}}
\textcolor{black}{To demonstrate the practical feasibility of transferring NAMO-LLM plans from simulation to real-world execution, we validated the method on a TurtleBot3 Waffle Pi with a 4-DOF OpenManipulator-X, operating in a 3D Mini-City platform; see Fig. \ref{pf2} and the supplemental material.
The robot performed state estimation, navigation, and manipulation using off-the-shelf ROS packages.
The experiment involved $n=19$ movable obstacles (traffic cones, construction blocks) and several fixed objects (buildings, pedestrian islands), with a minimal horizon $H_{\min}=2$.
To facilitate computation of free-space components $\ccalN(t)$ we over-approximated the robot as a disk of radius $r=15cm$ and the obstacles as polygons.
Our algorithm found the first feasible path in $1.14$ secs on average and the optimal $H_{\min}$ plan in $7.32$ secs.
}

\vspace{-0.1cm}
\section{Conclusion}
\vspace{-0.1cm}
\textcolor{black}{This paper proposed a probabilistically complete planner for NAMO problems. Comparative experiments verified its efficiency in cluttered environments. Future work will focus on extending to settings with unknown obstacle location and movability, and developing variants that avoid explicit C-space computation to better accommodate complex geometries and high-dimensional configuration spaces.
}
\vspace{-0.1cm}
\bibliographystyle{IEEEtran}
\bibliography{ref}

\end{document}